\documentclass{article}

\usepackage[a4paper, left=3cm, right=3cm, top=2.5cm, bottom=3cm]{geometry}

\usepackage{fancyhdr}
\fancypagestyle{plain}{
	\fancyhead[C]{\textit{Published in IEEE International Symposium on Biomedical Imaging (ISBI) 2026 Proceedings}}
	
	\fancyfoot[C]{\thepage}}

\usepackage{amsmath,graphicx}

\usepackage{enumitem}
\setlist{nosep, leftmargin=14pt}

\usepackage{graphicx,verbatim}
\usepackage{booktabs}
\usepackage[percent]{overpic}

\usepackage{todonotes}

\title{Uncertainty-guided Generation of Dark-field Radiographs}
\author{
\shortstack[c]{Lina Felsner$^{1,2,3}$, 
		Henriette Bast$^{4,5,6}$, 
		Tina Dorosti$^{4,5,6}$, 
		Florian Schaff$^{4,5,6}$, \\
		Franz Pfeiffer$^{4,5,6,7}$, 
		Daniela Pfeiffer$^{5,6,7}$, 
		Julia Schnabel$^{1,2,3,8}$}}
	
\date{}   

\begin{document}
	
%
\maketitle
\noindent
$^1$ School of Computation, Information and Technology, Technical University of Munich, 85748 Garching, Germany\\
$^2$ Munich Center for Machine Learning (MCML), Munich, Germany\\
$^3$ Institute of Machine Learning in Biomedical Imaging, Helmholtz Munich, 85764 Neuherberg, Germany\\
$^4$ Department of Physics, School of Natural Sciences, Technical University of Munich, 85748 Garching, Germany\\
$^5$ Munich Institute of Biomedical Engineering, Technical University of Munich, 85748 Garching, Germany\\
$^6$ Institute for Diagnostic and Interventional Radiology, School of Medicine and Health, TUM University Hospital Klinikum rechts der Isar, Technical University of Munich, 81675 Munich, Germany\\
$^7$ Institute for Advanced Study, Technical University of Munich, 85748 Garching, Germany\\
$^8$ School of Biomedical Engineering \& Imaging Sciences, King’s College, London, UK
	
	\vspace*{1cm}

\begin{abstract}
	X-ray dark-field radiography provides complementary diagnostic information to conventional attenuation imaging by visualizing microstructural tissue changes through small-angle scattering. However, the limited availability of such data poses challenges for developing robust deep learning models. 
	In this work, we present the first framework for generating dark-field images directly from standard attenuation chest X-rays using an Uncertainty-Guided Progressive Generative Adversarial Network. 
	The model incorporates both aleatoric and epistemic uncertainty to improve interpretability and reliability. 
	Experiments demonstrate high structural fidelity of the generated images, with consistent improvement of quantitative metrics across stages. 
	Furthermore, out-of-distribution evaluation confirms that the proposed model generalizes well. 
	Our results indicate that uncertainty-guided generative modeling enables realistic dark-field image synthesis and provides a reliable foundation for future clinical applications.
\end{abstract}


\noindent
\section{Introduction}
\label{sec:intro}

\begin{figure}[t]
  \begin{overpic}[width=\linewidth]{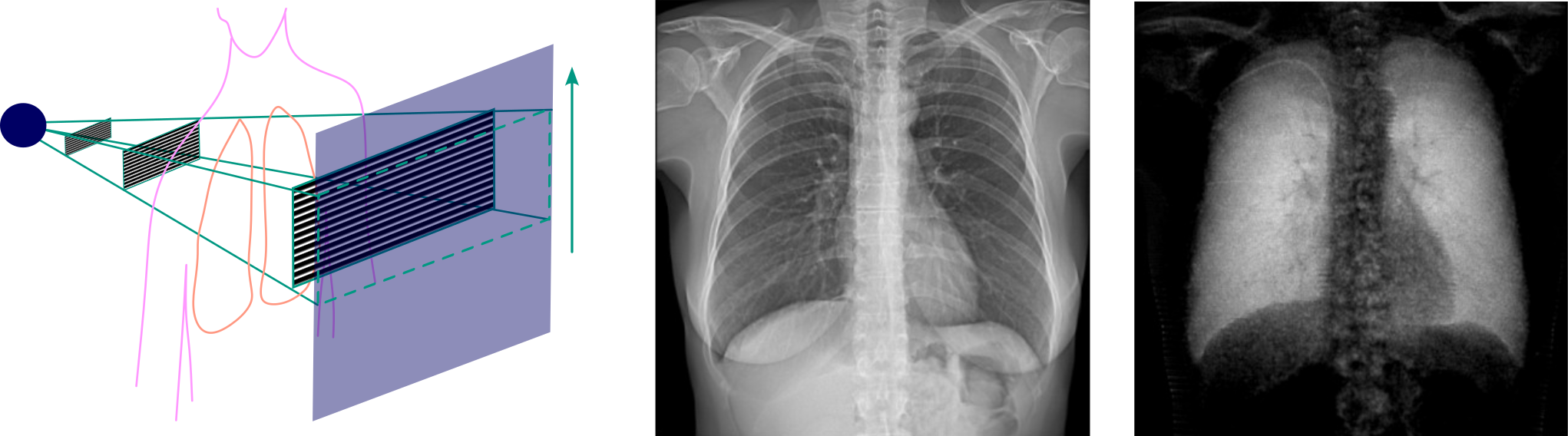}
	\put(0,23.5){\color{black}a}
	\put(43.0,23.5){\color{white}b}
	\put(74,23.5){\color{white}c}
\end{overpic}
	\caption{
		\textbf{(a)} Measurement setup for the grating-based dark-field scanner. The subject stands in the scanner while the grating assembly scans from bottom to top.
		\textbf{(b)} Attenuation image and \textbf{(c)} dark-field image of a healthy subject, jointly acquired in a single scan.%
		}
	\label{fig:setup}
\end{figure}

X-ray dark-field radiography is an advanced imaging modality that complements conventional X-ray by revealing microstructural features via small-angle scattering~\cite{Pfeiffer2008, Yashiro2010}. 
The Dark-field signal can be measured jointly with an attenuation image via grating-based interferometer setups~\cite{Pfeiffer2008,Willer2021} (see Fig.~\ref{fig:setup}).
Unlike standard attenuation-based radiography, where dense structures such as ribs and the heart dominate, dark-field contrast originates from microscopic interfaces that scatter X-rays.
The numerous alveolar air–tissue interfaces in healthy lungs produce a strong dark-field signal making healthy lungs appear bright, while diseases that affect alveoli (e.g. emphysema) cause patchy dark regions due to loss of scattering~\cite{Willer2021,Urban2023}.
Early clinical studies indeed suggest that dark-field chest radiography provides unique diagnostic value to quantify pulmonary emphysema in COPD patients~\cite{Willer2021, Urban2023} and enhance COVID-19 diagnosis~\cite{Gassert2025}.

Deep learning has shown transformative potential in medical imaging, achieving strong performance in tasks such as nodule detection, disease classification, and image segmentation on conventional radiographs~\cite{zhou2021review}.
However, robust training of such models typically demands large, diverse datasets, a requirement that is challenging for a new modality like dark-field, where only limited human data currently exist.
One promising strategy to mitigate this problem is synthetic data generation: using generative models to create realistic artificial images for data augmentation~\cite{Chen2017,Yi2019,Armanious1806,Upadhyay2021,nafi2024diffusion}.
By augmenting the training set with high-fidelity simulated samples~\cite{zhou2021review}, or using synthetic pretraining~\cite{moroianu2025improvingperformancerobustnessfairness} model performance can be improved.

Generative Adversarial Networks (GANs) have been especially influential because they can synthesize realistic images via adversarial training without explicit density modeling, and tend to require fewer training samples than diffusion models.
Many GAN architectures have been explored for medical imaging tasks, from MedGAN~\cite{Armanious1806}, a general-purpose medical image translation framework operating end-to-end on image inputs,
to task-specific models like PathologyGAN~\cite{ClaudioQuiros2020}, which learns to generate and analyze realistic histopathology images of cancer tissue.
Overall, the medical imaging field is increasingly leveraging such synthetic data techniques to supplement limited datasets and improve AI model training~\cite{fok2024adversarial}.

An essential aspect for the clinical deployment of generative models is the quantification of uncertainty. 
Uncertainty generally falls into two categories: aleatoric uncertainty, which reflects noise inherent to the observed data, and epistemic uncertainty, representing the uncertainty arising from the model itself—this latter type can be reduced with sufficient data~\cite{kendall2017uncertainties}.
Uncertainty quantification techniques are critical for mitigating uncertainties in both optimization and decision-making processes~\cite{Abdar2021}. 
Recent research indicates that incorporating uncertainty estimation not only enhances interpretability but can also lead to improved model performance~\cite{Upadhyay2021}.
For instance, Upadhyay~\textit{et al.}~\cite{Upadhyay2021} introduced an uncertainty-guided GAN approach for image translation trained in a cascaded matter.
They show that the visual quality and quantitative metrics improve across these refinement phases, and that the uncertainty maps help the model focus on areas of poor synthesis.
Moreover, the uncertainty estimation improves interpretability and may support expert assessment and safe deployment in medical contexts~\cite{Upadhyay2021}.
These approaches highlight that developing reliable medical image synthesis systems requires not only generating realistic images but also identifying low-confidence regions to mitigate false certainty.

In this paper, we propose to use an uncertainty-guided image synthesis framework to generate X-ray dark-field images from standard attenuation chest X-rays. 
To our knowledge, this is the first attempt to translate conventional radiographs into dark-field contrast via a generative model, enabling the large-scale generation of virtual dark-field data from widely available chest X-rays.
The main contributions of this work can be summarized as follows:
\begin{enumerate}
	\item To the best of our knowledge we are the first to generate \textit{dark-field images} from conventional 2D X-ray radiographs.
	\item We explicitly model both \textit{aleatoric} and \textit{epistemic} uncertainty within the image generation process to improve interpretability and reliability.
	\item We evaluate the proposed model on the \textit{ChestX-ray} dataset to assess its robustness and generalization under domain-shift conditions.
\end{enumerate}
The proposed framework not only provides a new tool for generating synthetic dark-field datasets to facilitate machine learning, but also moves toward reliable AI-assisted interpretation by quantifying uncertainty in the generated images. 
Together, these contributions aim to accelerate the use of dark-field imaging for lung disease diagnosis by combining image generation with principled uncertainty estimation.

\section{Methods}

\begin{figure}[t]
	\includegraphics[width=\linewidth]{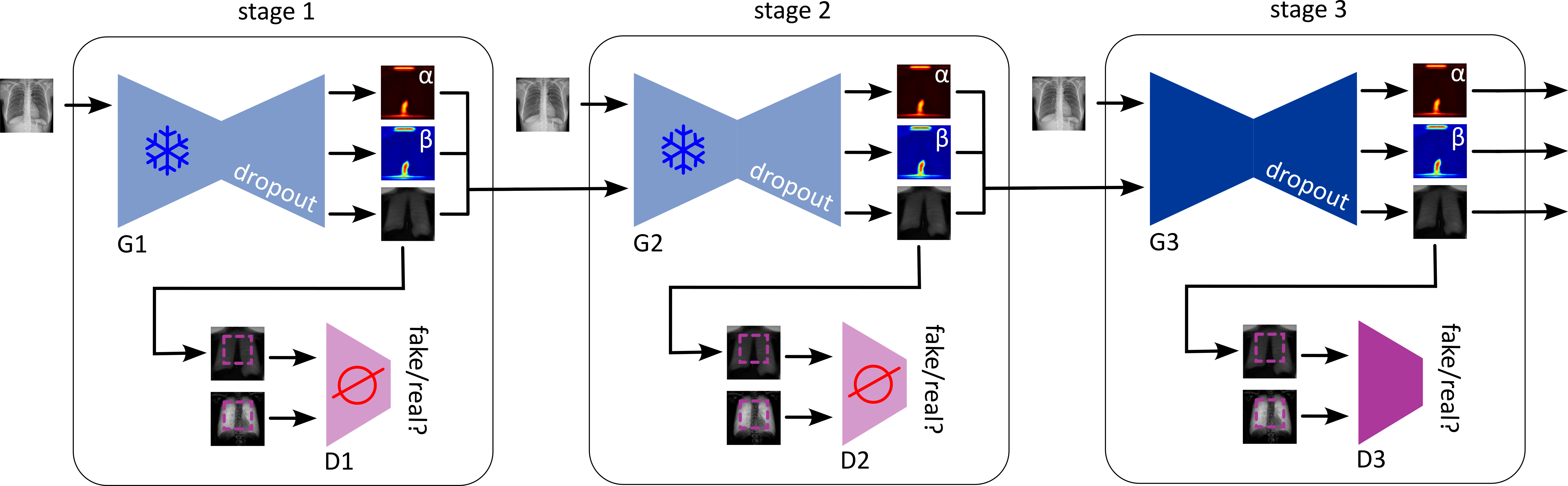}
	\caption{
		Proposed Uncertainty-Guided Progressive GAN framework, illustrated for model training at stage three. 
		At this stage, the previous layers are frozen. The aleatoric uncertainty estimates are used as attention maps to guide the model's weights for refinement in the subsequent stage. 
		Dropout is incorporated to the generators at each stage.%
	}
	\label{fig:model}
\end{figure}

We propose a generative approach to synthesize dark-field contrast images from conventional attenuation chest radiographs using a GAN-based framework that incorporates both aleatoric and epistemic uncertainty. 
Our method builds on the Uncertainty-Guided Progressive GAN framework introduced by Upadhyay~\textit{et al.}~\cite{Upadhyay2021}, based on a progressive learning scheme and uses aleatoric uncertainty estimates as attention maps to guide the models weight for refinement.
In this work, we explicitly model both forms of uncertainty:
(i) Aleatoric uncertainty, which accounts for the inherent noise in the observations, and 
(ii) Epistemic uncertainty, which arises from limited knowledge about the model and is addressed through Monte Carlo sampling via dropout at inference time.

To model aleatoric uncertainty, the Uncertainty-Guided Progressive GAN~\cite{Upadhyay2021} introduces two pixel-wise parameters, $\alpha$ and $\beta$, which define a generalized Gaussian distribution rather than assuming a fixed Gaussian or Laplacian noise model. Specifically, for each pixel $(i, j)$, the network predicts $\alpha_{ij}$ (the scale parameter) and $\beta_{ij}$ (the shape parameter). The scale parameter $\alpha$ controls the spread of the distribution—larger values correspond to broader distributions and hence higher uncertainty, while smaller values indicate narrower, more confident predictions. The shape parameter $\beta$ determines the tail behavior: $\beta = 2$ corresponds to a Gaussian distribution, $\beta = 1$ to a Laplacian distribution, and $\beta < 1$ produces heavier tails, allowing the model to capture non-Gaussian noise and outliers.
Together, $\alpha$ and $\beta$ enable the network to learn pixel-wise uncertainty. 
The standard deviation (effective uncertainty) for a generalized Gaussian can then be expressed as:
\begin{equation}
	\sigma = \alpha \sqrt{ \frac{ \Gamma\left( \frac{3}{\beta} \right) }{ \Gamma\left( \frac{1}{\beta} \right) } }\, ,
\end{equation}
where $\Gamma(\cdot)$ is the Gamma function. 

To model epistemic uncertainty, we employ Monte Carlo dropout during inference, following the approach proposed by Gal and Ghahramani~\cite{gal2016dropout}. 
In this framework, dropout layers remain active at test time, and multiple stochastic forward passes are performed through the network. 
Those repeated stochastic passes approximate a posterior over the network weights, capturing the model uncertainty and ultimately reflecting how confident the network is in its learned parameters.

To better represent the characteristic noise structure of dark-field images, we added a residual consistency loss. 
Residuals are computed by subtracting a locally blurred version (box blur) from both the predicted and target images, and their L1 difference is penalized to encourage realistic noise and texture,

\begin{figure}[t]
	{\footnotesize 
	\hspace{0.9cm} Attenuation \hspace{1.2cm} Real dark-field \hspace{0.8cm} Generated dark-field}\\[2pt]
	\rotatebox{90}{\hspace{.95cm}\footnotesize Patient 1}  
	\includegraphics[width=\linewidth, trim={0 40 0 40}, clip]{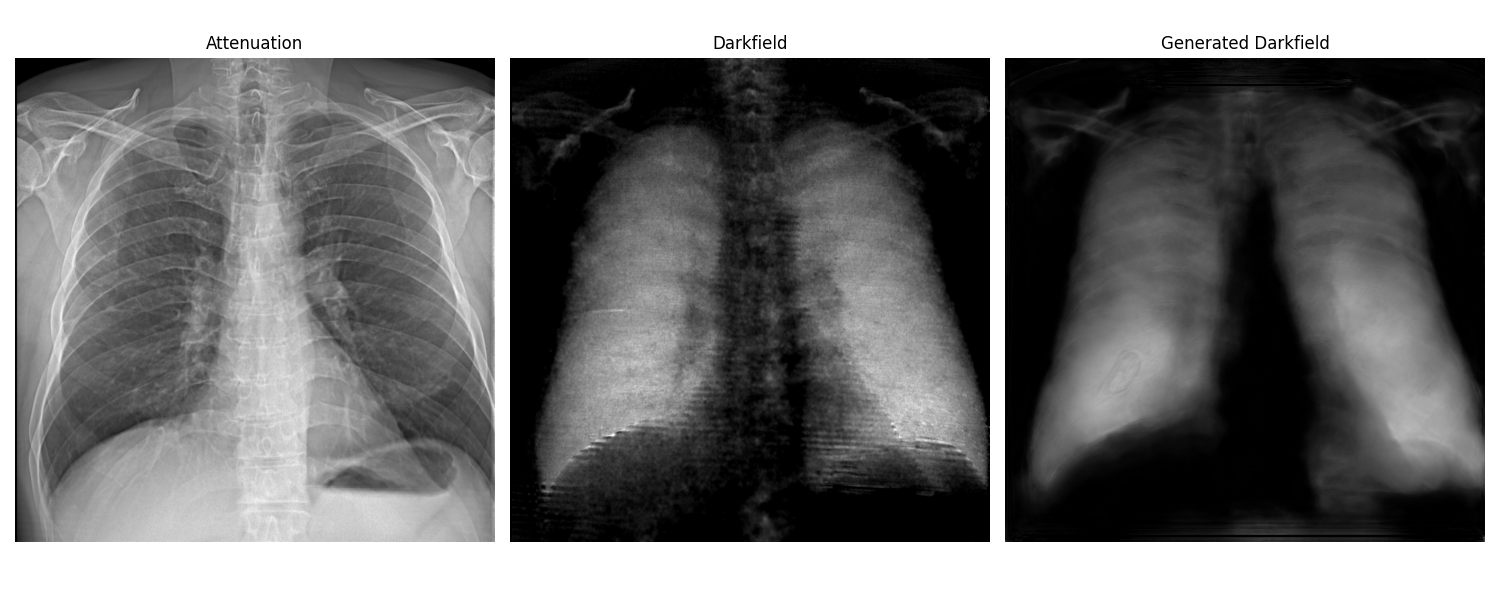}
	\rotatebox{90}{\hspace{.95cm}\footnotesize Patient 2}  
	\includegraphics[width=\linewidth, trim={0 40 0 40}, clip]{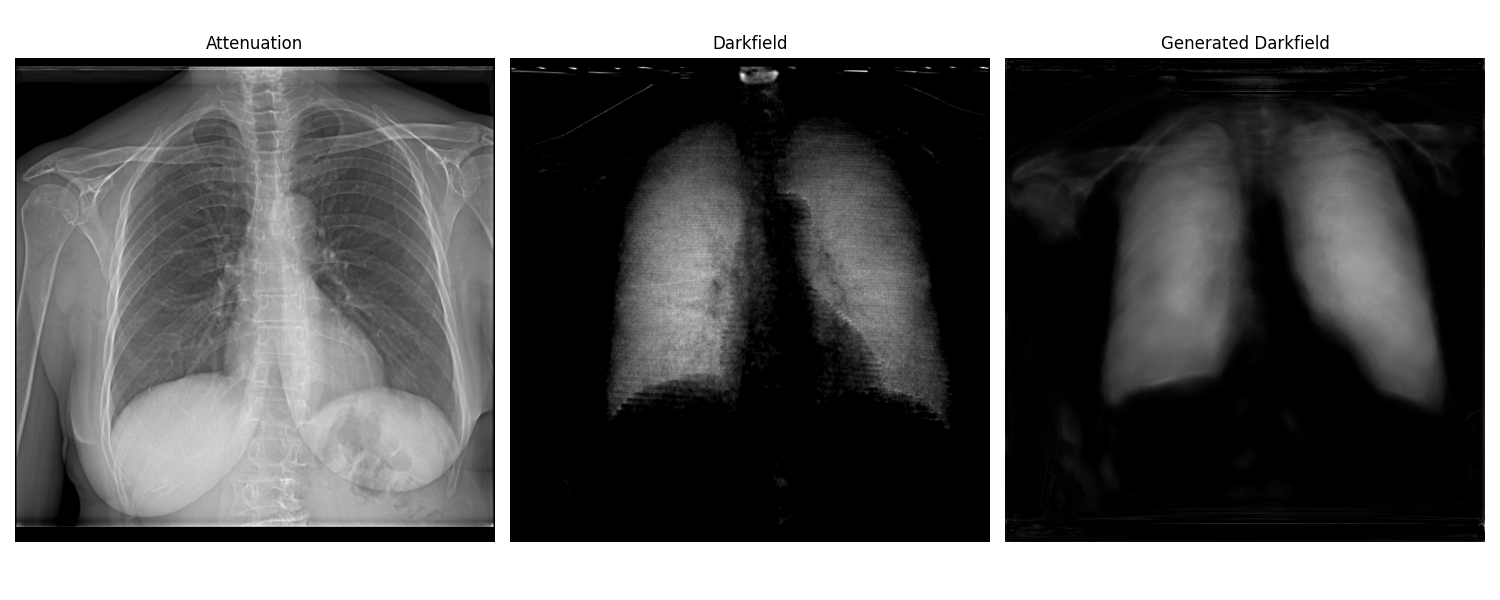}
	\caption{Comparison of attenuation, real dark-field, and generated dark-field images for two patients.
	}
	\label{fig:gen}
\end{figure}

\section{Experiments}

\noindent \textbf{Dark-field Data:~}
We used an in-house dataset from Klinikum rechts der Isar consisting of paired chest radiographs and corresponding dark-field images from 269 patients during inhalation. 
The images are jointly acquired in a grating-based scanner, as shown in Fig.~\ref{fig:setup}, resulting in perfect pose alignment between the attenuation radiograph and the dark-field image.
The data were split into training (227), validation (15), and test (27) subsets.
We applied data augmentation (random rotations, flips, and contrast jitter) during training to improve robustness and reduce overfitting.

\vspace*{0.2cm}
\noindent \textbf{Training:~}
Similar to Upadhyay~\textit{et al.}~\cite{Upadhyay2021} we used a progressive learning scheme with three stages.
We trained the model using the Adam optimizer with a learning rate of $8e-6$ and lambda parameters $0.8$ and $0.001$ for all stages. 
Models were trained for $50$ epochs each, with a cosine annealing learning rate scheduler. 
Dropout was applied during both training and inference, using a dropout rate of $0.1$ and generating $20$ samples per image during testing.
Training was performed on a single NVIDIA RTX A6000 GPU using PyTorch 2.2.

\vspace*{0.2cm}
\noindent \textbf{Evaluation Metrics:~}
Performance was quantitatively evaluated using the Mean Squared Error~(MSE), Structural Similarity Index~(SSIM), and Peak Signal-to-Noise Ratio~(PSNR) to assess both structural fidelity and perceptual image quality.

\vspace*{0.2cm}
\noindent \textbf{Uncertainty Analysis:}~
In the first experiment, we evaluate the model on the dark-field data to assess the overall quality of the generated images. 
Both quantitative metrics and qualitative evaluations are performed to comprehensively assess the model’s performance and the interpretability of its uncertainty maps.
We report the results for all three model stages to also illustrate the progressive refinement.

\vspace*{0.2cm}
\noindent \textbf{Out-of-Domain Generalization:~}
In a second experiment, we evaluate the generalizability of the proposed model using an external NIH Chest X-ray dataset~\cite{summers2019nih}. 
The data was downsampled to $947\times956$ to fit the dark-field data.
The goal of this experiment is to investigate how well the model and its uncertainty estimation mechanisms generalize to conventional chest X-ray data, in contrast to the grating-based setup used for the dark-field data.
In particular, we focus on the behavior of the uncertainty maps and whether they indicate higher uncertainty in unfamiliar anatomical regions or image characteristics.

\section{Results}

Figure~\ref{fig:gen} shows the attenuation image, real dark-field and generated dark-field image for two patients. 
We find a good visual agreement between the generated dark-field and real measured dark-field. 
We particularly note pronounced differences in dark-field intensity across regions and patients. 
The model preserves these contrasts, which is critical for visualizing microstructural variation. 
We attribute the sharper appearance of the original dark-field images to additional scattering contributions that become visible through the dark-field signal reconstruction process. 
In particular, the horizontal stripe artifacts originating from the image formation process are not well captured by our model.
Despite the remaining limitations, the model achieves a strong overall correspondence with real dark-field images.

The agreement between the real and generated dark-field is confirmed by the quantitative vales presented in table~\ref{tab:metrics_summary}.
\begin{table}[t]
	\centering
	\footnotesize 
	\caption{Mean and standard deviation of Mean Squared Error (MSE), Peak Signal-to-Noise Ratio (PSNR), and Structural Similarity Index (SSIM) for each model stage between the real and generated dark-field image.}
	\label{tab:metrics_summary}
	\begin{tabular}{lccc}
		\toprule
		Stage  		& MSE 				& PSNR 					& SSIM \\
		\midrule
	 1  &$	0.0131\pm	0.0067	$ & $19.35\pm	2.14$ & $	0.38\pm	0.06$ \\
	 2  &	$0.0125\pm	0.0066	$ & $19.57\pm	2.24$ & $	0.47\pm	0.05$ \\
	 3  &$	0.0123\pm	0.0067$   & $19.71\pm	2.37$ & $	0.52\pm	0.05$ \\
		\bottomrule
	\end{tabular}
\end{table}
For all three metrics(MSE, PSNR, SSIM), we observe a consistent improvement across the model stages, indicating that the progressive refinement leads to higher image fidelity and better structural consistency.
The final stage achieves the lowest MSE and highest SSIM and PSNR values, confirming that the model successfully reconstructs fine structural details.

\begin{figure}[t]
	{\footnotesize 
  	\hspace{0.3cm} Dark-field \hspace{0.975cm} $\alpha$ \hspace{1.35cm} $\beta$ \hspace{0.95cm} Aleatoric \hspace{0.43cm}  Epistemic}\\[2pt]
	\rotatebox{90}{\hspace{0.1cm}\footnotesize Stage 1}  
	\includegraphics[width=\linewidth, trim={0 560 210 60}, clip]{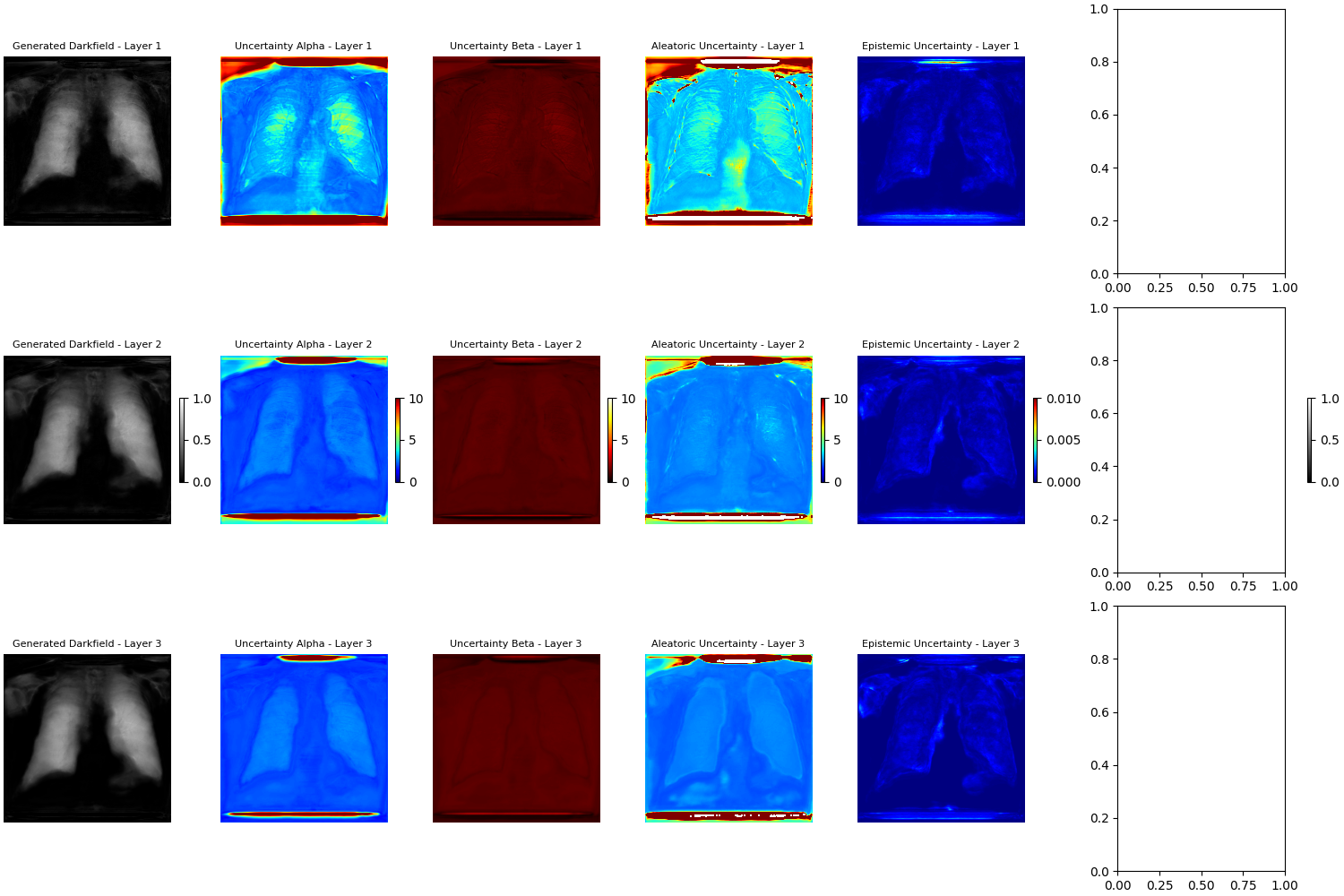}
	\rotatebox{90}{\hspace{0.1cm}\footnotesize Stage 2}  
	\includegraphics[width=\linewidth, trim={0 315 210 300}, clip]{figures/plot_gen_alpha_beta_var_test3.png}
	\rotatebox{90}{\hspace{0.1cm}\footnotesize Stage 3}  
	\includegraphics[width=\linewidth, trim={0 80 210 540}, clip]{figures/plot_gen_alpha_beta_var_test3.png}
	\caption{Results and corresponding uncertainty estimates across the three training stages for a single patient. All images are cropped to the lung region to enhance visualization.}
	\label{fig:uncertainty}
\end{figure}
Figure~\ref{fig:uncertainty} shows the generated dark-field image along with the predicted $\alpha$ and $\beta$ values, the overall aleatoric uncertainty, and the epistemic uncertainty for all three model stages for a representative patient.
For the aleatoric uncertainty, we observe a general decrease with increasing model stage, indicating improved confidence in the image reconstruction. 
In the final stage, the highest uncertainty is concentrated within the lung regions themselves.
In contrast, the epistemic uncertainty remains relatively constant across stages, suggesting stable model behavior after training convergence.

\begin{figure}[t]
	{\footnotesize 
	\hspace{0.6cm} Attenuation \hspace{.45cm} Gen. dark-field \hspace{0.75cm} Aleatoric \hspace{0.85cm}  Epistemic}\\[2pt]
	\rotatebox{90}{\hspace{.65cm}\footnotesize Patient 1}  
	\includegraphics[width=\linewidth, trim={0 230 0 230}, clip]{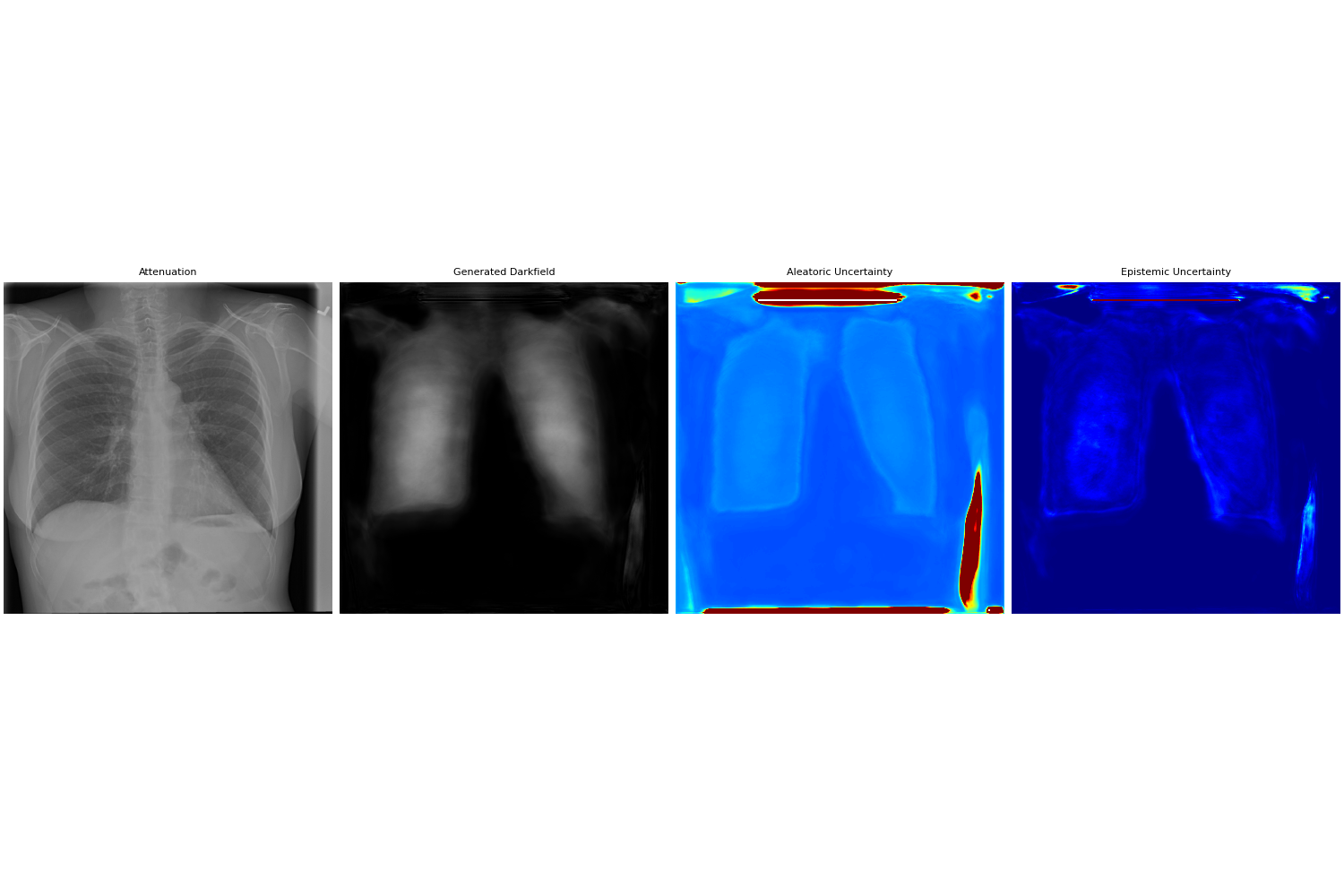}
	\rotatebox{90}{\hspace{.6cm}\footnotesize Patient 2}  
	\includegraphics[width=\linewidth, trim={0 230 0 230}, clip]{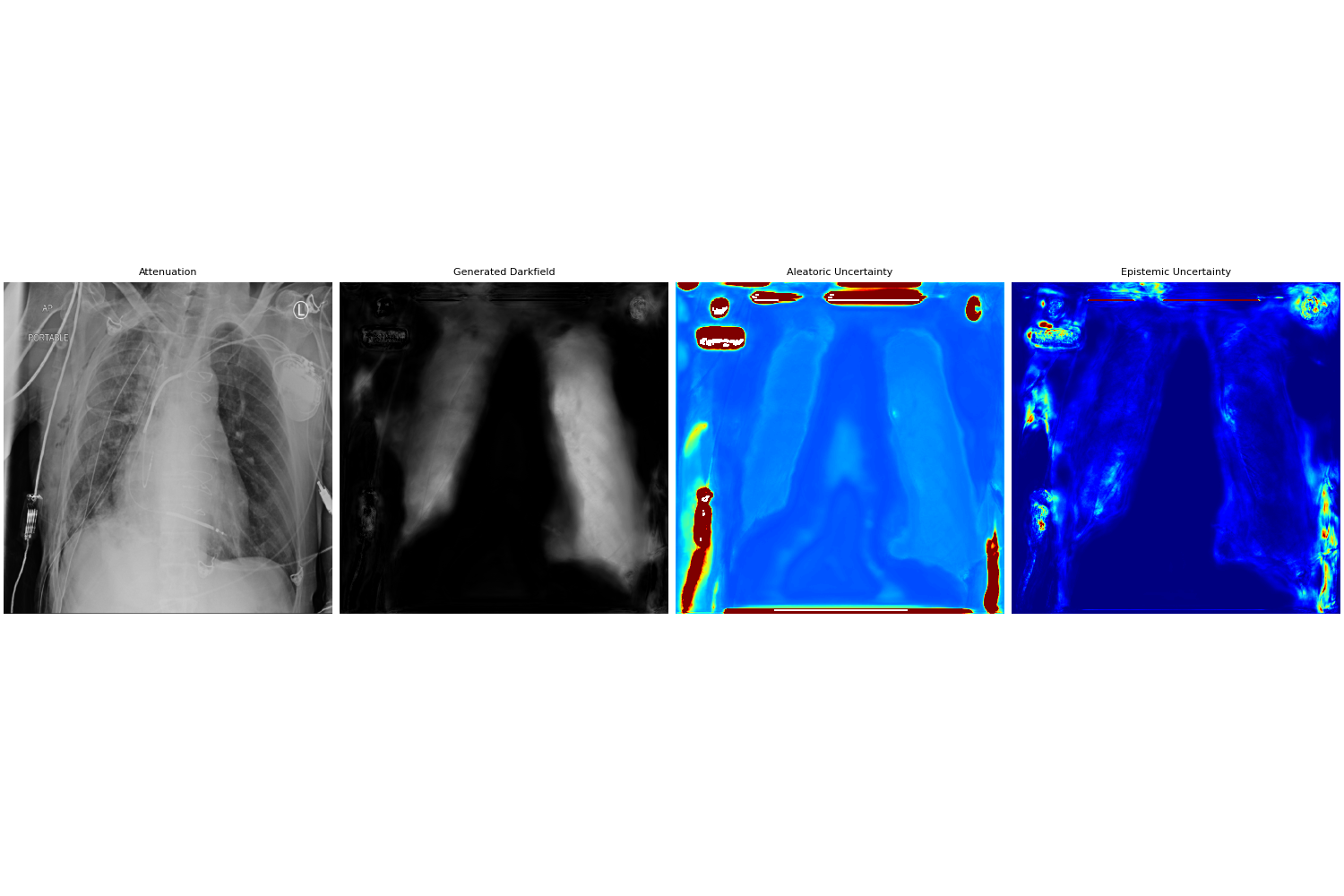}
	\rotatebox{90}{\hspace{.55cm}\footnotesize Patient 3}  
	\includegraphics[width=\linewidth, trim={0 230 0 230}, clip]{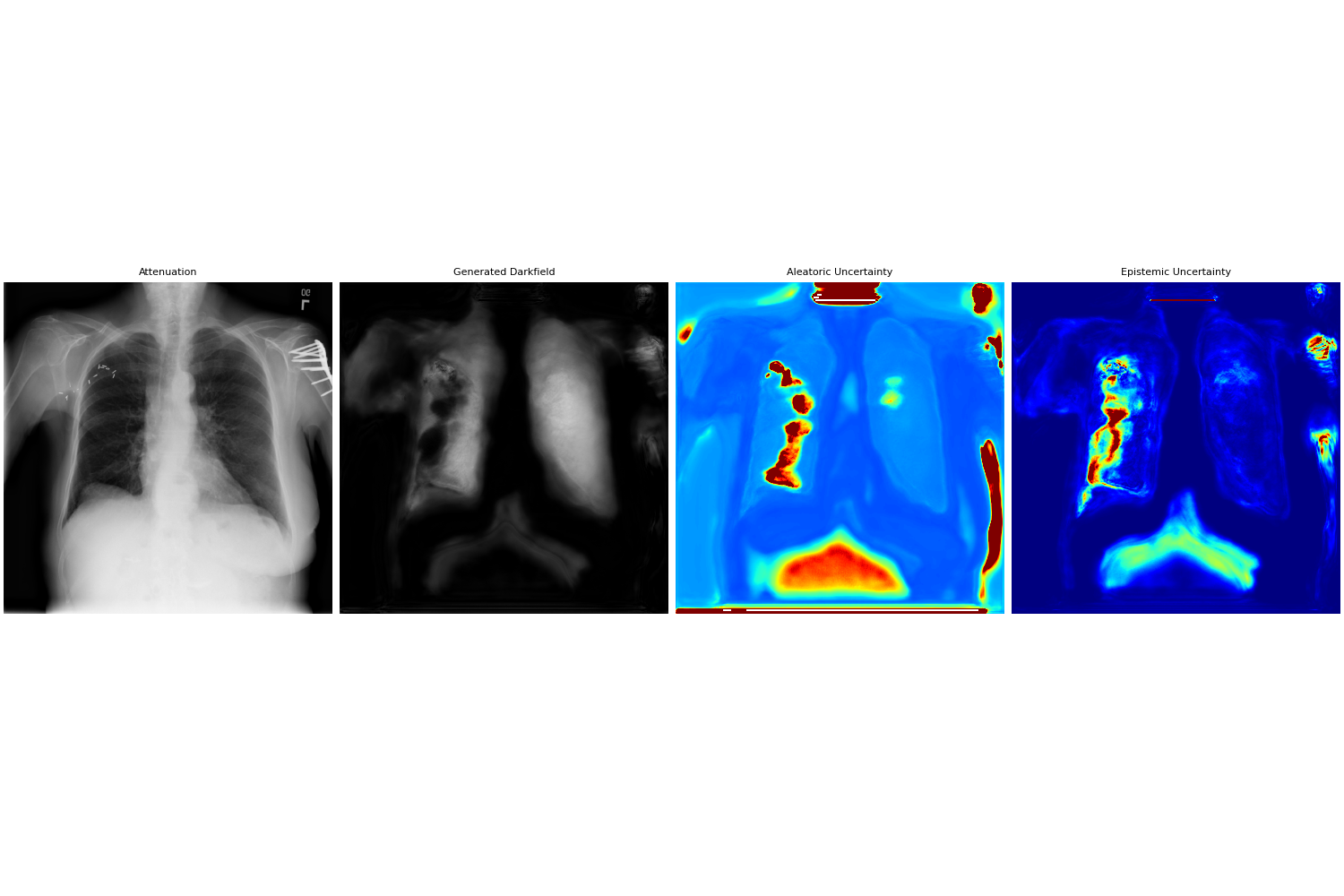}
	\caption{Attenuation and generated dark-field images along with their corresponding uncertainty estimates for the out-of-distribution NIH Chest X-ray dataset, shown for 3 patients.}
	\label{fig:ood}
\end{figure}

Figure~\ref{fig:ood} shows the results for the out-of-distribution data from the NIH Chest X-ray dataset. We present the attenuation images, the generated dark-field images, and the corresponding aleatoric and epistemic uncertainty maps for two patients.
For the first patient, the model generates realistic dark-field images, with both uncertainty types remaining at similar and relatively low levels.
The attenuation image of the second patient contains cables, a pacemaker, and lead markers, features that are not or only rarely present in the training data. Despite these out-of-distribution elements, the model is able to generate a visually plausible dark-field image. The corresponding uncertainty estimates reflect this deviation from the training domain.
For the third patient, however, we observe a failure case: the left lung exhibits artifacts in the generated dark-field image. This issue is also reflected in increased aleatoric and epistemic uncertainty. In addition, the model shows pronounced uncertainty in the upper abdominal region.

\section{Discussion and Conclusion}

We presented the first deep learning–based framework for generating dark-field images from attenuation data, integrating both aleatoric and epistemic uncertainty estimation.
The results demonstrate that the proposed progressive model can accurately reconstruct dark-field images from attenuation data, achieving high image fidelity and structural consistency. 
Quantitatively, all metrics show consistent improvement across the model stages, confirming that the progressive refinement effectively enhances fine structural detail and reduces artifacts. 
The final stage achieves the best overall performance, showing that the model learns to recover increasingly realistic and anatomically consistent dark-field representations.
Although the SSIM value of $0.5$ indicates only moderate structural similarity, this is expected given the cross-modality nature of the task and inherent differences between the attenuation and dark-field domains. 
Moreover, GANs are known to suppress stochastic noise~\cite{durall2020watch}, which may explain both the absence of horizontal stripe artifacts in the generated outputs and the resulting lower SSIM.

The uncertainty analysis provides insight into the model’s behavior. 
As artifacts such as rib shadows diminish across the refinement stages, the aleatoric uncertainty also decreases, suggesting that it serves as a reliable indicator of reconstruction quality. 
Higher uncertainty values are observed in the lung regions for all three stages. 
This is expected, as lung tissue exhibits noisy and heterogeneous structures, making the prediction more ambiguous. 
The model thus appears to meaningfully capture both data-driven and structural uncertainty.

The out-of-domain evaluation further supports the robustness of the approach. Despite the presence of unseen image features such as cables, pacemakers, and lead markers, the model produces high-fidelity dark-field images, while also expressing appropriate uncertainty levels. This behavior reflects well-calibrated uncertainty estimation and provides confidence that the model can flag unreliable predictions when confronted with unfamiliar anatomical or device structures. 
This is further illustrated by the third patient, where the overall image contrast is notably higher than in other cases, likely contributing to the model’s failure and the corresponding increase in both aleatoric and epistemic uncertainty.

Although the aleatoric and epistemic uncertainty maps show similar spatial patterns, they reflect different aspects of model confidence.
High aleatoric uncertainty marks regions with ambiguous signals, while elevated epistemic uncertainty highlights areas where the model may not generalize well.
Together, they offer a more complete view of model reliability and data quality.

Future work will focus on extending the model to 
clinically relevant variations. 
Specifically, 
it would be valuable to generate dark-field signals in relation to pulmonary disease severity, 
as the contrast is expected to vary with pathological changes. 
Incorporating a parameterized framework to 
model varying degrees of 
noise and structural alterations could further enhance the model’s utility for data generation, enabling more robust training of downstream tasks.
Additionally, diffusion-based generative models could be explored, though this direction remains limited by available dataset size.
Future work could also include comparisons with alternative generative architectures to 
clarify performance differences across model classes.
Finally, the presented framework could generate synthetic dark-field datasets for data augmentation and supporting downstream tasks such as disease classification.

\section{Acknowledgments}
\label{sec:acknowledgments}
This work is supported by the Konrad Zuse School of Excellence in Reliable AI (relAI).
We acknowledge financial support through the European Research Council (ERC Synergy Grant SmartX, SyG 101167328), the Center for Advanced Laser Applications (CALA), and the Free State of Bavaria under the Excellence Strategy of the Federal Government, as well as by the Technical University of Munich – Institute for Advanced Study.
%

\bibliographystyle{plain}
\bibliography{lit}

\end{document}